\documentclass{article}

\usepackage{arxiv}

\usepackage[utf8]{inputenc} 
\usepackage[T1]{fontenc}    
\usepackage{hyperref}       
\usepackage{url}            
\usepackage{booktabs}       
\usepackage{amsmath,amsfonts}       
\usepackage{nicefrac}       
\usepackage{microtype}      
\usepackage{lipsum}		
\usepackage{graphicx}
\usepackage[sort]{natbib}
\usepackage{doi}
\usepackage{xcolor}

\title{Explainable Unsupervised Anomaly Detection with Random Forest}

\author{
	Joshua S. Harvey\\
	Prospect 33, LLC, USA\\
	New York, NY 10006 \\
	\texttt{joshua.harvey@prospect33.com} \\
	\And
	Joshua Rosaler \\
	BlackRock, Inc, USA\\
	New York, NY 10001 \\
	\texttt{joshua.rosaler@blackrock.com} \\
	\And
	Mingshu Li \\
	Prospect 33, LLC, USA\\
	New York, NY 10006 \\
	\texttt{mingshu.li@prospect33.com} \\
	\And
	Dhruv Desai \\
	BlackRock, Inc, USA \\
	New York, NY 10001 \\
	\texttt{dhruvdesai@alumni.upenn.edu} \\
	\And
	Dhagash Mehta \\
	BlackRock, Inc, USA\\
	New York, NY 10001 \\
	\texttt{dhagash.mehta@blackrock.com} \\
}

\renewcommand{\shorttitle}{Explainable Unsupervised RF Anomaly Detection}

\hypersetup{
pdftitle={Explainable Unsupervised Anomaly Detection with Random Forest},
pdfsubject={cs.LG, stat.ML},
pdfauthor={Joshua S.~Harvey, Joshua Rosaler, Mingshu Li, Dhruv Desai, Dhagash Mehta},
pdfkeywords={Anomaly detection, Random Forest, Explainable Machine Learning},
}

\begin{document}
\maketitle

\begin{abstract}
	We describe the use of an unsupervised Random Forest for similarity learning and improved unsupervised anomaly detection. By training a Random Forest to discriminate between real data and synthetic data sampled from a uniform distribution over the real data bounds, a distance measure is obtained that anisometrically transforms the data, expanding distances at the boundary of the data manifold. We show that using distances recovered from this transformation improves the accuracy of unsupervised anomaly detection, compared to other commonly used detectors, demonstrated over a large number of benchmark datasets. As well as improved performance, this method has advantages over other unsupervised anomaly detection methods, including minimal requirements for data preprocessing, native handling of missing data, and potential for visualizations. By relating outlier scores to partitions of the Random Forest, we develop a method for locally explainable anomaly predictions in terms of feature importance.

\end{abstract}


\section{Introduction}
Ensuring data quality is the first required step upstream of all data-driven functions. Integral to this is the detection of anomalies in datasets, records with such great dissimilarity from the bulk of the data, that they are best explained as having been generated by some other, potentially erroneous, process. Detecting anomalies is therefore critical for both rectifying faulty data input processes, and identifying signals of interest in veridical data.

Popular anomaly detection algorithms typically require extensive, hands-on preprocessing of datasets to be successful. This is because ultimately such methods, whether they be distance-based such as \textit{k}-nearest neighbor (kNN), or density-based such as kernel density estimation (KDE), assume either implicitly or explicitly some distance metric over the data. As such, they are sensitive to preprocessing operations such as the scaling of numerical features and the encoding of categorical features. Particularly in contexts where a high volume of diverse datasets must be absorbed, such preprocessing can present a bottleneck, requiring dataset-specific subject matter expertise.

Fundamentally, anomalies are those data points that show markedly low similarity to the rest of the dataset, and a critical aspect underpinning anomaly detection techniques is the ability to measure, or learn to measure, the similarity or dissimilarity between data points. Some notion of similarity is integral to all learning and discriminating systems \citep{tversky}, including recommendation systems, image recognition, and anomaly detection. Yet there are many potential ways to calculate similarity, which may be more or less useful depending on the task at hand. \emph{Similarity learning} is a subfield within machine learning that optimizes distance functions, the computing of similarity between observations as a function of their features \citep{santini}, towards achieving some goal.

One approach to computing similarity is to employ a machine learning model to learn a distance function (which may not be a distance metric in the strict sense) over a dataset, so-called \textit{similarity learning}. While this presents some added computational demands for data quality pipelines, it obviates the need for arbitrary dataset-specific preprocessing decisions, allowing the pipeline to be fully automated. Random Forests present an attractive candidate model for this task, due to their versatility in handling high-dimensional datasets of mixed variable types. Depending on their implementation, they may also have native support for missing data as well as un-encoded categorical features. As noted by Breiman and Cutler, distance (or proximity) discovery is ``one of the most useful tools in Random Forests'' \citep{breiman-cutler-blog}.

While typically used according to a supervised learning paradigm, where some feature of interest in the dataset serves as the target for classification or regression, Random Forests can also be used in an unsupervised manner. This may be achieved either by completely randomizing the splitting decision function (extremely randomized trees, or \textit{ExtraTrees}) \citep{Geurts2006}, or by training the Random Forest to discriminate between real, original data and synthetic data generated to retain some but not all statistical properties of the original \citep{shi}. Such approaches have previously been explored for their application to anomaly detection \citep{mensi2022rf,PUGGINI2015583,rhodes,baron2016weirdest}, but have not been directly compared.

In this paper, we examine the distance measure learned by a Random Forest trained to discriminate between original data and synthetic data sampled uniformly over the bounds of the original ($RF_{uni}$). We show that while the \textit{ExtraTrees} distance function approximates Euclidean distances over numerical features, $RF_{uni}$ learns a representation that exaggerates distances at the boundary of the data manifold between outliers and inliers, making it particularly attractive for anomaly detection. Over a large number of anomaly detection benchmark datasets \citep{adbench}, we show its superior performance over the use of distances obtained from the \textit{ExtraTrees} model, and highly favorable performance compared to other state-of-the-art unsupervised anomaly detection algorithms. We also show how outlier predictions can be related to feature importance of the Random Forest, for locally explainable anomaly detection.

\section{Previous work and our approach}

Our approach can be reduced to four steps: (i) train a Random Forest with unsupervised learning on the dataset; (ii) compute distances from the trained Random Forest; (iii) compute outlier scores from distances; and (iv) evaluate anomaly detection performance. Each of these steps has multiple potential methods and has been the subject of significant research effort. In this section, we review this background and describe our chosen approach.

\subsection{Unsupervised Learning with Random Forests}

Typically, random forests are trained to predict a particular target variable within the data. While one can calculate distances from a random forest trained in this supervised manner, the model will only learn to represent relationships in the data that can be exploited to predict the target variable. This may be helpful, but in the limiting case (where a target variable shows complete conditional independence over all combinations of all other variables) could be completely uninformative. An alternative approach, again originally described by Breiman and Cutler, is to train a random forest to discriminate between the real dataset and some synthetic data, generated to share some but not all of statistical properties of the real data \citep{breiman-cutler-blog}.

Shi and Horvath investigated the use of unsupervised Random Forest for similarity learning on datasets without inherent class structure \citep{shi}. They identified two different methods for generating the artificial data, which the Random Forest model must learn to discriminate from the real data. The first is that proposed by Breiman and Cutler, of sampling each feature from its univariate distribution in the real data, in effect shuffling each feature independently. This method, originally called `Addcl1', creates a synthetic dataset that retains the marginal distributions of the real data, but not the joint distributions, and has been the most widely adopted approach for implementing unsupervised Random Forests \citep{madhyastha2019geodesic,auret2010unsupervised,seligson2005global}.

The second method was to sample each feature from a uniform distribution over the bounds of the real data. Considerably less attention has been paid to this method, originally called `Addcl2', although we argue it has attractive properties for anomaly detection. While Addcl1 only learns similarity structure that is predictive of the joint distribution of the data, anomalies may affect features that exhibit no conditional dependence on other features, and Addcl1-determined transformations may not be sensitive to such deviations. Addcl2, on the other hand, will be sensitive to any features whose distributions differ from that of a uniform distribution. We also consider the limiting the case for each method. For Addcl1, this is when no conditional independence is present for any combinations of features in the data. The resultant Random Forest then partitions the space proportionally to the density of the data, in effect transforming the space to one approximating even density, which is deleterious to anomaly detection.

The limiting case for Addcl2 occurs when all features approximate uniform distributions. In this case, the Random Forest partitions would be drawn from a uniform distribution, and therefore is equivalent to the \emph{ExtraTrees} model \citep{Geurts2006}, learning a mapping that approximates Euclidean distances. We therefore apply the Addcl2 method, referring to unsupervised Random Forest models built with this approach as $RF_{uni}$.

\subsection{Random Forest Distances}

Random Forest, as an adaptive nearest neighbor algorithm, offers a robust solution for local distance learning across disparate datasets from diverse domains. Central to this approach is first the fitting of a random forest model, optimized for a particular task, followed by the calculation of pairwise distances from that model, represented in an $N\times{N}$ distance matrix, given $N$ observations in the dataset. Many methods have been proposed for calculating pairwise distances from the leaf indices of a Random Forest, the original being simply the proportion of trees in which points do not coterminate, regardless of in-bag or out-of-bag status \citep{breiman}. While attractive in its simplicity, the distances recovered from a Random Forest with this method do not match up with its predictive performance. Instead, ``geometry and accuracy preserving'' (GAP) proximities can be computed by differentially weighting in-bag and out-of-bag observations when tallying their colocations to the model's leaves \citep{rhodes2023geometry}:

\begin{equation} \label{GAPprox}
p^{GAP}_{i,j} = \frac{1}{|S_{i}|} \sum_{t \in S_{i}} \frac{c_j(t)I[j \in J_{i}(t)]}{|M_{i}(t)|}
\end{equation}

\noindent where $S_{i}$ is the set of trees in the RF for which observation $i$ is out of bag, $M_i(t)$ is the multiset 
\footnote{Recall that a multiset is a generalization of the concept of a set, allowing for repetition among the elements of the set, where the number of repetitions of a unique element in the multiset is known as its multiplicity.} of bagged points in the same leaf as $i$ in tree $t$, $J_{i}(t)$ is the corresponding set (i.e., without repetitions) of bagged points in the same leaf as $i$ in tree $t$, and $c_{j}(t)$ is the multiplicity of the index $j$ in the bootstrap sample. 

From GAP proximities, we can then compute symmetrized GAP distances as such:

\begin{equation} \label{GAPdist}
    d^{GAP}_{i,j} =
    \begin{cases}
        0, & \text{if } i = j \\
        (0.5 \cdot (p^{GAP}_{i,j} + p^{GAP}_{j,i}))^{-1}, & \text{otherwise}
    \end{cases}
\end{equation}

With GAP distances, it is possible to exactly reconstruct the predictions of the Random Forest as weighted averages of training target labels. This makes it a particularly natural, effective, and interpretable notion of distance from Random Forest representations (as given by leaf indices). While other more complex distance calculations have been developed, such as those considering not only leaf colocations but wider notions of path similarity, these have been found to have little impact on anomaly detection performance for $ExtraTrees$ models \citep{mensi2022rf}. As such, we use the GAP method for calculating RF distances.

\subsection{Computing outlier scores from distances}

There are many possible ways to compute outlier scores from inter-point distances\footnote{Besides distance-based anomaly detectors, any other method can be applied (e.g. density-based detectors) after first embedding distances in Euclidean space. This may be done with Principal Coordinate Analysis \citep{gower1966} (metric multidimensional scaling), or other embedding methods.}. Common approaches include variants of kNN, such as the distance to the \textit{k}-th neighbor of each point, or the average distance to the \textit{k} nearest neighbors \citep{Ramaswamy2000}. While often performing well, such methods require setting the hyperparameter of \textit{k}. Our tests showed that the performance of these methods on Random Forest distances is highly sensitive to the choice of \textit{k}, and that the optimal value of \textit{k} varies significantly between datasets.

A method that requires no hyperparameter tuning is the use of the median distance to all other points as a measure of centrality. One drawback of this method is its sensitivity to the presence and proportion of outliers in the dataset. To mitigate this, we propose to first identify the most central 50\% of the data, and then only consider distances to these central points. Given a dataset, $X=\{x_1, x_2, \ldots, x_n\}$ of $n$ points, we find the subset of observations with the lowest median GAP distances to all other observations:

\[
X_{\mathrm{central}} = \{ x_i \in X : \mathrm{Rank}(d_{\mathrm{med}}^{GAP}(x_i)) \leq \lfloor 0.5n \rfloor \}.
\]

Where $d_{med}^{GAP}(x_i)$ is the median GAP Random Forest distance for observation $x_i$ to all other observations. For each observation we then compute its outlier score, $O(x_i)$ as the median GAP distance to observations in \( X_{\mathrm{central}} \):
\[
O(x_i) = \mathrm{med} \left( \{ d_{i,j}^{GAP} : j \in X_{\mathrm{central}} \} \right).
\]

\subsection{Evaluating the performance of anomaly detectors}

Anomaly detection is a central field of research within applied statistics and machine learning, with various techniques developed to identify unusual instances in data. For a comprehensive treatment of the evolution and methodologies in anomaly detection, including robust statistical methods and modern machine learning approaches, readers are encouraged to consult the meta-survey by Olteanu et al. \citep{olteanu}. Recent research has focused on deep learning-based outlier detection techniques (e.g. DeepAnT framework for time-series data \citep{munirDeepAnT}) and various hybrid models that combine deep learning with traditional machine learning techniques to enhance anomaly detection capabilities, e.g. \citep{munirFuseAD}.

A notable contribution to this field includes the development of libraries such as PyOD (Python Outlier Detection) \citep{pyod}, an open-source Python library specifically designed for detecting outliers in multivariate data. PyOD includes a comprehensive collection of anomaly detection algorithms (e.g. kNN, Isolation Forest, autoencoders etc.). ADBench (Anomaly Detection Benchmark) \citep{adbench} is another significant contribution to the field of anomaly detection. It provides a standardized set of benchmark datasets for evaluating the performance of different anomaly detection algorithms. ADBench includes a diverse set of real-world datasets from numerous domains, each with ground-truth labels of whether observations are outliers, facilitating the evaluation of anomaly detectors as binary classifiers via methods such as the receiver operating characteristic (ROC) curve.

ADBench contains 47 datasets for benchmarking anomaly detection algorithms, gathered across diverse domains such as healthcare, finance, sociology, image processing, and the physical sciences. The advantage of using ADBench lies in immediate access to multiple, harmonized datasets, which have already been extensively tested and evaluated for anomaly detection. However, it also presents drawbacks. Documentation for datasets can be cumbersome to locate, such that it can be difficult to trace back to their original reference. More importantly, the datasets have already been preprocessed, with numerical values being scaled and categorical features being encoded. Therefore, testing on ADBench datasets is not completely representative of end-to-end anomaly detection in the real world. One of the key benefits of the RF distance-based method described here is the lack of required preprocessing for diverse datasets, particularly with respect to native handling of null values and outliers; this benefit will not be demonstrated by using datasets that are already preprocessed.

As there is no universally applicable definition of what constitutes an anomaly, the labeling of ground truth anomalies in datasets is somewhat subjective. In several of the ADBench datasets, the `anomaly' subset of data actually represents a distinct class within the dataset, with observations of this class being highly similar to one another. One example is the \textit{optdigits} dataset, consisting of 5,216 8-by-8 pixel images of hand-drawn digits. In this dataset, observations are labeled as anomalous if they depict a zero, one of the 10 digits featured. We argue it does not make sense to consider such a class as anomalous---although it is a small fraction of the dataset, such observations are to be expected, and are in no sense `deviations' from the majority of the data, which are themselves divided amongst the other nine categories of digit. Similarly, other datasets may be strongly enriched for `anomalous' observations, such as medical datasets containing a proportion of pathological samples far higher than would be expected in the general population. While one approach would be to train models only on inlier points (the approach taken in \citep{mensi2022rf}), such an approach cannot be considered truly unsupervised, requiring anomaly labels for training. For such reasons, we excluded nine of the 47 real-world datasets from experiments (see App. \ref{sec:exclusions}).

For all experiments, we randomly sampled 1,000 observations from each dataset five times, preserving the balance of inliers and outliers. Datasets with fewer than 1,000 observations were excluded, leaving 26 remaining datasets. Smaller datasets pose problems for evaluating anomaly detection performance, as metrics such as AUCROC can be overly optimistic for limited test set sizes \citep{novello2024outofdistributiondetectionuseconformal}. When evaluated on datasets excluded from the main analysis, we found no significant differences in the performance of different anomaly detection algorithms, with the exception of OCSVM performing worse than kNN ($p=0.0218$).

\section{Results}
\subsection{Visualizing Unsupervised Random Forest Transformations}\label{sec:visualizing}

\begin{figure}[!ht]
	\centering
	\includegraphics[width=1\textwidth]{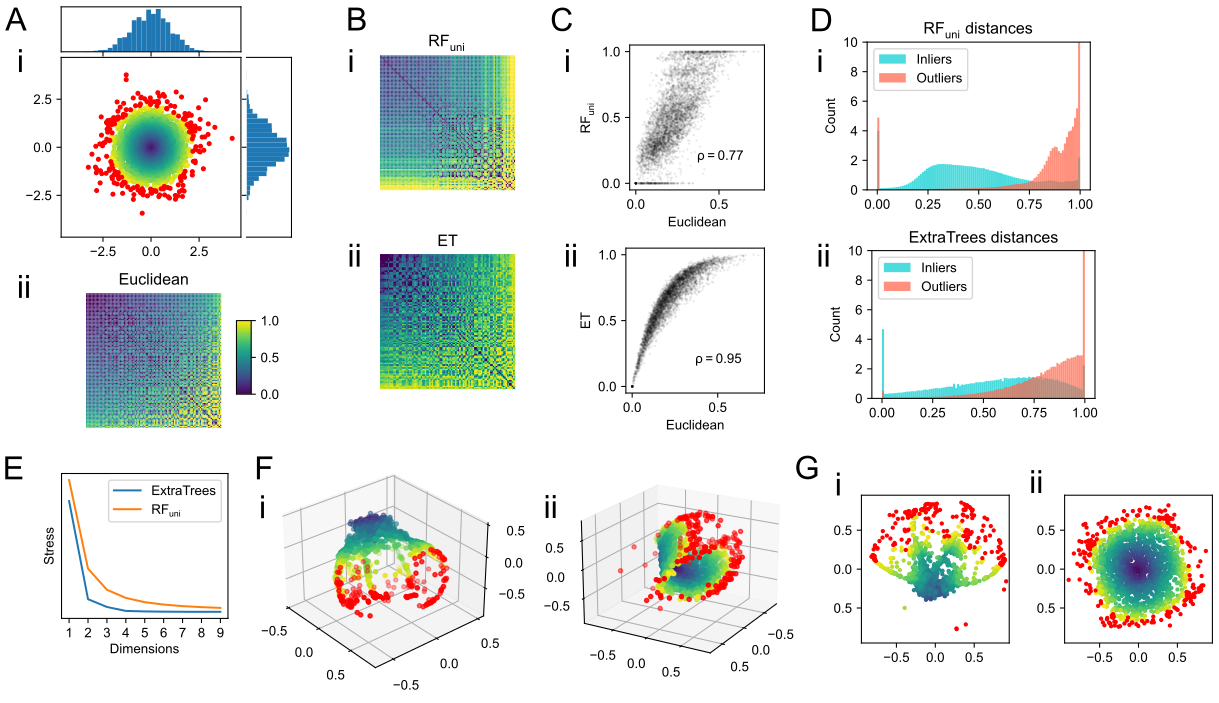}
	\caption{\textbf{Embedding two-dimensional Gaussian data with unsupervised RF distances.} A) i) Data simulated from a two-dimensional Gaussian distribution. Points with distances from the origin above the 90th percentile are colored red (outliers). ii) A matrix of Euclidean distances, sorted by each point's distance from the origin. B) Distance matrices for Random Forest GAP distances obtained from the (i) $RF_{uni}$ and (ii) ${ExtraTrees}$ models. C) Spearman rank correlation, $\rho$, between distances in measurement space (Euclidean) and RF distances. D) Histograms of inter-point RF distances for inliers (blue) and outliers (red). E) Stress plot for multidimensional scaling (MDS) of RF distances. F) MDS embedding in three dimensions for (i) $RF_{uni}$ and (ii) ${ExtraTrees}$ RF distances. G) MDS embedding in two dimensions for RF distances.}
	\label{Simulated_data}
	\end{figure}

As there are multiple ways of training a Random Forest for unsupervised learning on a dataset, it is of value to understand what different approaches achieve, and how they transform the data from measurement space to a fitted model's representation. For the purposes of visualization, we explored this with simulated data, drawn from a two-dimensional Gaussian distribution, with points in the 90th percentile of distances from the distribution center designated as outliers (Fig. \ref{Simulated_data}.a.i). A distance matrix showing inter-point Euclidean distances is shown in Figure \ref{Simulated_data}.a.ii, with points sorted by their distance from the origin. We then fitted two Random Forest models, one tasked with discriminating between the real data and synthetic data drawn from a uniform distribution over its bounds ($RF_{uni}$), and one using a completely randomized splitting function ($ExtraTrees$). From each fitted model, we then computed GAP distances (Fig. \ref{Simulated_data}.b).

Of note, the $RF_{uni}$ model significantly reshapes the inter-point distance relationships in the data, giving GAP distances with a Spearman rank correlation $\rho=0.77$ to the original Euclidean distances in measurement space (Fig. \ref{Simulated_data}.c.i). The $ExtraTrees$ model, on the other hand, largely preserves the distance relationships of the data ($\rho=0.95$, Fig. \ref{Simulated_data}.c.ii). Histograms of inter-point distances reveal that the $RF_{uni}$ model tends to exaggerate distances for outlier points, while the distances for inlier points tend to decrease, with respect to the $ExtraTrees$ model (Fig. \ref{Simulated_data}.d). This increasing isolation of outlier points suggests the approach may be useful for outlier detection.

To visualize these transformations, we embedded GAP distances with multidimensional scaling (MDS) \citep{gower1966}, which attempts to accommodate inter-point distances in a low-dimensional embedding with minimal distortion (`stress'). While the $ExtraTrees$ distances could be accommodated with little stress in two dimensions, reflecting the original measurement space of the data, $RF_{uni}$ distances could not (Figure \ref{Simulated_data}.e).

Visualizing three-dimensional MDS embeddings, it can be seen that $RF_{uni}$ applies a significant, anisometric reshaping of the data, with innermost points being pulled closer together and outermost points being pushed away through a higher dimension from the more central data (Figure \ref{Simulated_data}.f). This is in contrast to the $ExtraTrees$ model, which approximately retains the two-dimensional input space distances of the data, albeit with some curvature (Figure \ref{Simulated_data}.b.ii).

Two-dimensional embedding with MDS shows again how the $RF_{uni}$ model pushes outlier points to the outer reaches of the mapping while drawing inliers closer together (Figure \ref{Simulated_data}.g.i). The $ExtraTrees$ model, on the other hand, does not distort the mapping of points from their measurement space (Figure \ref{Simulated_data}.g.ii).

\subsection{Evaluating Anomaly Detection Performance}\label{sec:evaluating}

\begin{figure}[!ht]
	\centering
	\includegraphics[width=1\textwidth]{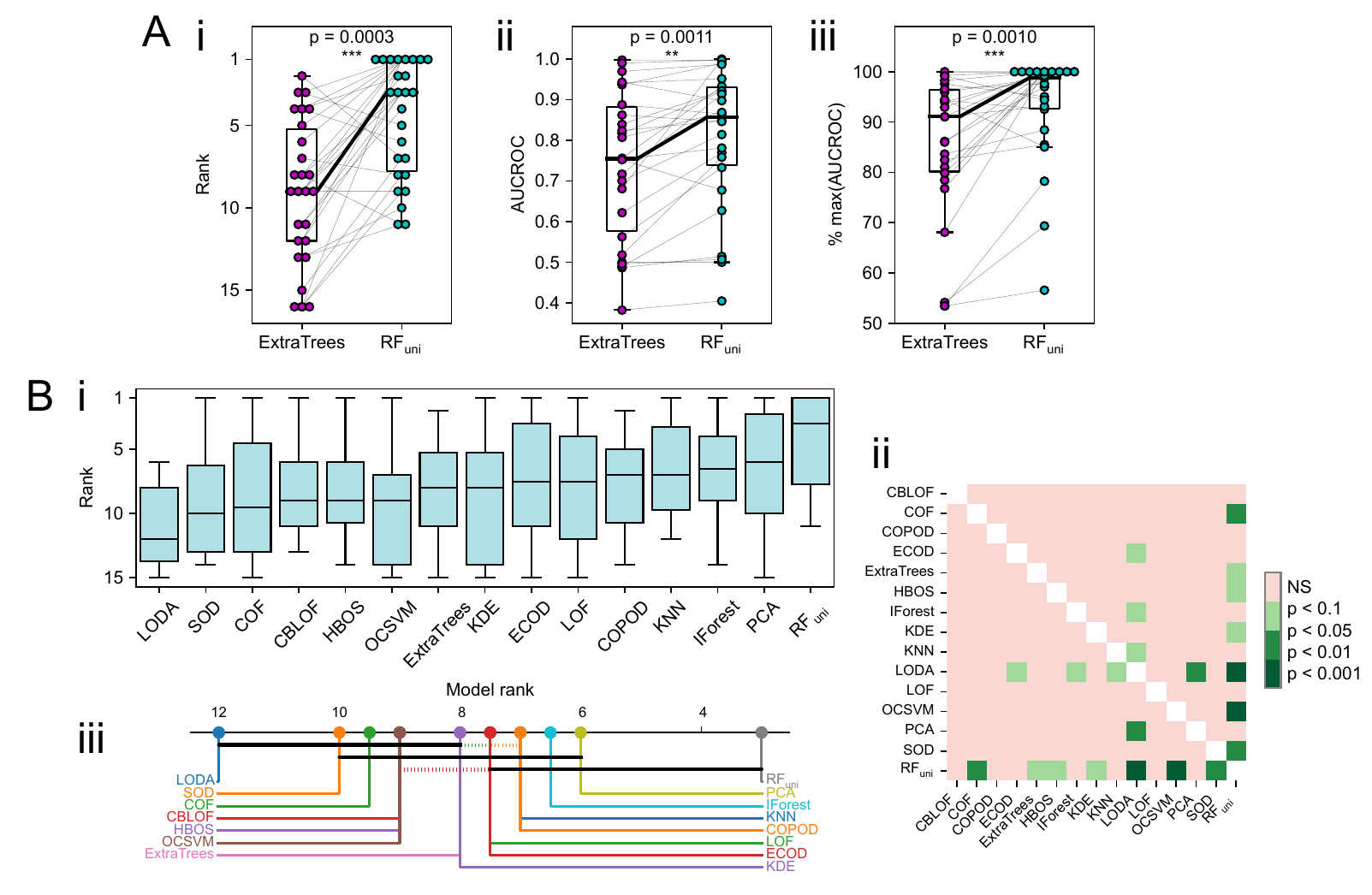}
	\caption{\textbf{Anomaly detection performance on benchmark datasets.} A) Direct comparison between anomaly detectors distances from either the \textit{ExtraTrees} or $RF_{uni}$ Random Forest model, aggregated across benchmark datasets. Comparisons of i) ranking amongst other unsupervised detectors, ii) AUCROC score, and iii) percent of AUCROC score achieved by best performing detector for each dataset. $p$-values indicate result of a Wilcoxon signed-rank test. B) Results for all unsupervised anomaly detectors. i) Boxplots of ranked performance for each detector, sorted left to right by superior performance. ii) Pairwise comparisons computed with a Conover post-hoc test, with p-values adjusted for multiple comparisons via the Holm–Bonferroni method. iii) A critical difference diagram connecting detectors without significant differences at $\alpha=0.1$. Dashed colored lines indicate no critical difference between specific detectors, despite critical differences between their intermediately ranked detectors.}
	\label{ADbench_results}
	\end{figure}

The tendency of $RF_{uni}$ to exaggerate the isolation of outlier points makes it an attractive candidate for use in anomaly detection. To test this, we first compared the performance directly of detectors using either $RF_{uni}$ or $ExtraTrees$ distances, aggregated across the 26 ADBench datasets we included in our analysis. We found a significant improvement in anomaly detection performance when using $RF_{uni}$ distances, when evaluated according to the ranking of the detector (measured against other unsupervised detectors\footnote{We evaluated the performance of the following unsupervised detectors: OCSVM \citep{Scholkopf1999}, KNN \citep{Ramaswamy2000}, LOF \citep{Breunig2000}, COF \citep{Tang2002}, PCA \citep{Shyu2003}, CBLOF \citep{He2003}, IForest \citep{Liu2008}, KDE \citep{Latecki2007}, SOD \citep{Kriegel2009}, HBOS \citep{Goldstein2012}, LODA \citep{Pevny2016}, COPOD \citep{Li2020}, and ECOD \citep{Li2023}.}), the AUCROC score, and the percentage of AUCROC score achieved by the detector as a percentage of the maximnum AUCROC score achieved by the best-performing detector for each dataset (Figure \ref{ADbench_results}.a).

We then compared the performance of $RF_{uni}$ against other popular unsupervised anomaly detectors, using the ADBench benchmark datasets (\ref{ADbench_results}.b). A boxplot of ranked accuracy shows little difference in the performance of detectors aggregated across all datasets, with the exception of $RF_{uni}$, which has a median performance ranking of three out of 15 (\ref{ADbench_results}.b.i). This is reflected in the findings of a Conover post-hoc test, which shows that while most unsupervised detector comparisons exhibit no statistically significant differences in performance, $RF_{uni}$ is significantly better than half of the other detectors (p<0.1, adjusted for multiple comparisons via the Holm-Bonferroni method) (\ref{ADbench_results}.b.ii). These comparisons are also represented in a critical difference diagram \citep{demsar2006statistical} (\ref{ADbench_results}.b.iii).

To confirm that the elevated performance of $RF_{uni}$ could not be attributed to our method of converting inter-point distances to outlier scores, we also applied this method on Euclidean distances. As expected, this detector performed similarly to the $ExtraTrees$ detector, confirming that the $RF_{uni}$ data transformation accounted for its superior anomaly detection performance over the benchmark datasets.

\subsection{Anomaly Detection with Missing Values}

\begin{figure}[!ht]
	\centering
	\includegraphics[width=.8\textwidth]{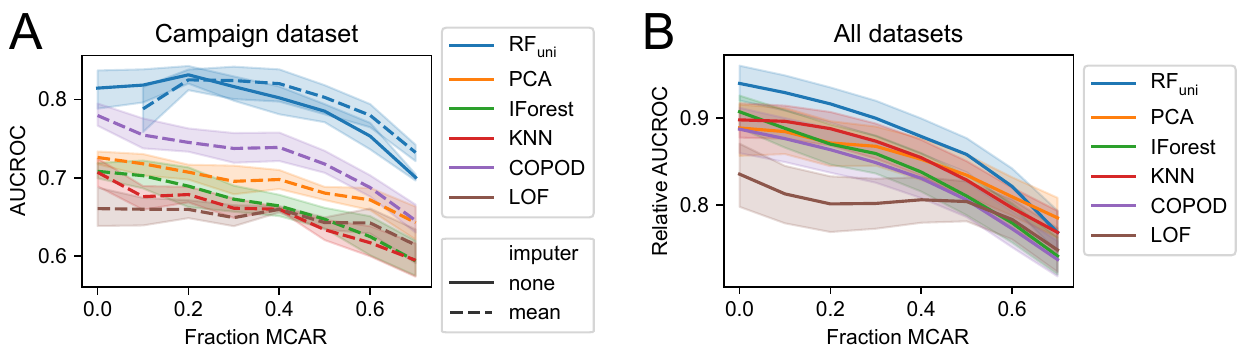}
	\caption{\textbf{Anomaly detection performance with missing data.} A) The performance of $RF_{uni}$ and other top-performing unsupervised anomaly detectors with missing data on the \emph{Campaign} dataset, evaluated with AUCROC. All detectors were tested on data imputed with the mean (dashed lines), while $RF_{uni}$ was also applied directly on missing data (solid line). Shaded area indicates standard deviation over 5 repeats. B) Aggregate performance across all datasets for varying levels of missingness, following mean imputation. AUCROC scores for each dataset are normalized to that of the best-performing detector on complete data.}
	\label{ADbench_results_missing}
	\end{figure}

Across many sectors and industries, datasets often contain a significant proportion of missing values. To evaluate the stability of unsupervised Random Forest anomaly detection on missing data, we tested its performance on benchmark datasets for varying levels of missingness in the data (missing completely at random, MCAR), and compared its stability over increasing levels of missingness to that of other detectors. For some datasets, $RF_{uni}$ maintains a superior performance without prior imputation of missing data (e.g. the \emph{Campaign} dataset, Fig. \ref{ADbench_results_missing}.a). However, we find that in general its performance is improved by first imputing missing values with the mean.

Aggregating results across all datasets, we find that $RF_{uni}$ maintains its competitive performance over other unsupervised detectors, up to missingness levels of 60\% (Fig. \ref{ADbench_results_missing}.b).

\subsection{Explainability}

While there are several model-agnostic explainability frameworks that can be leveraged for anomaly detectors (eg. SHAP \citep{Lundberg2017}, LIME \citep{Ribeiro2016}), here we explore a method that capitalizes on the explainable properties of Random Forest estimators. Our method is inspired by the popular approach of constructing saliency maps to visualize feature importance for image class prediction models \citep{simonyan2013deep}. While saliency maps visualize input features that impact model performance as a function of neural network gradients, here we develop an approach for counterfactual explanations, i.e. minimal differences in inputs that impact model predictions \citep{Guidotti2024}.

Given a dataset, $X$, the $RF_{uni}$ outlier score of each observation is given by $O(x_i)$. For each point $x_i$, we want to find $\nabla O(x_i)$, the gradient of $O$ at $x_i$. We first compute local gradients of the outlier score in measurement space, solving the least squares solution of $O$ over neighborhoods of the dataset, with $N_k(x_i)$ being the set of $k$ nearest neighbors of point $x_i$:

\[
A_i = \sum_{j \in N_k(x_i)} (x_j - x_i)(x_j - x_i)^T,
\]
\[
b_i = \sum_{j \in N_k(x_i)} (x_j - x_i)(O(x_j) - O(x_i)),
\]
\[
	\nabla O(x_i) = A_i^{-1}b_i
\]

Once the gradient field $\nabla O$ is computed, a trajectory can be charted given a learning rate, $l$:

\[
	p_i = x_i + l \cdot \nabla O(x_i),
\]
\[
x_{i+1} = \underset{x \in X}{\text{argmin}} \, ||x - p_i||_2
\]

\begin{figure}[!ht]
	\centering
	\includegraphics[width=.95\textwidth]{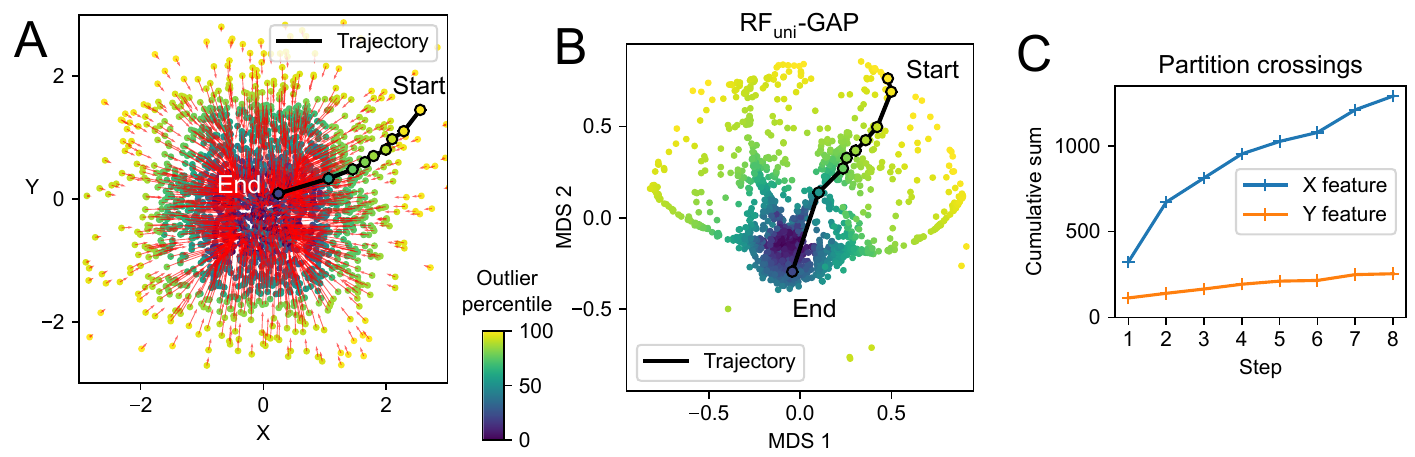}
	\caption{\textbf{Explainability of $RF_{uni}$ anomaly detector outlier scores.} A) A trajectory through the outlier gradient field on two-dimensional Gaussian data. Gradients are shown in red arrows. B) The same trajectory visualized in an MDS embedding of GAP distances computed from the $RF_{uni}$ outlier detection model. C) Cumulative tally of which $RF_{uni}$ partitions are crossed along the trajectory.}
	\label{Explainability_toy}
	\end{figure}

Such a trajectory can be interpreted as a sequence of counterfactual explanations, minimal changes in inputs that result in the largest possible impacts in outlier score. Figure \ref{Explainability_toy}.a shows the outlier gradient field computed over the two-dimensional toy dataset from Figure \ref{Simulated_data}, with an example counterfactual trajectory. The same trajectory is also shown in an MDS embedding of GAP distances obtained from the $RF_{uni}$ model (Fig. \ref{Explainability_toy}.b).

We then compute the feature importance of each step along the counterfactual trajectory, with respect to the $RF_{uni}$ model. This is accomplished by identifying intersections made by the straight line segment connecting $x_i$ and $x_{i+1}$, through the hyperrectangular partitions of the Random Forest (described in App. \ref{sec:intersections}). In this way, we are able to explain $RF_{uni}$ outlier scores with respect to an observation's feature values (Fig. \ref{Explainability_toy}.c).

\begin{figure}[!ht]
	\centering
	\includegraphics[width=1\textwidth]{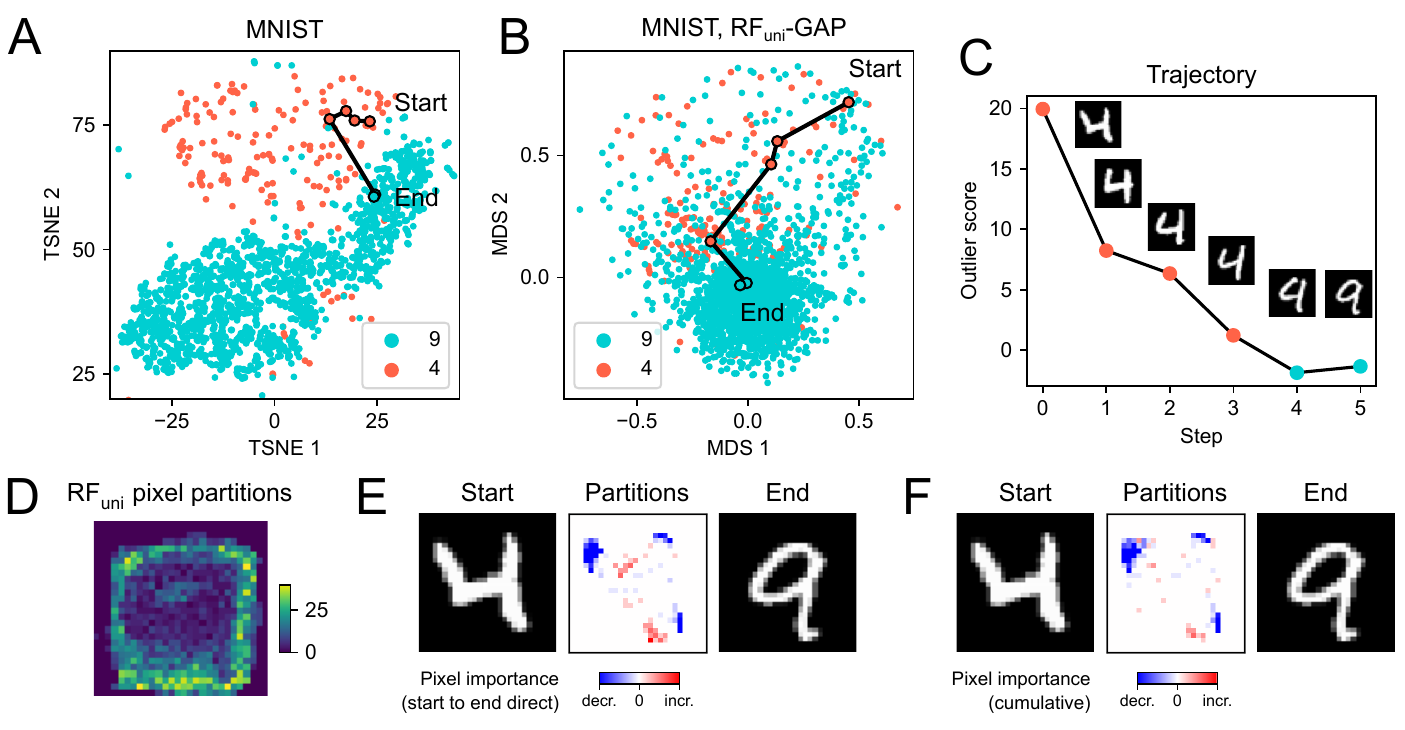}
	\caption{\textbf{Explainable anomaly detection with the MNIST dataset.} A) The trajectory from an anomaly (digit 4, red) to inliers (digit 9, blue) in the MNIST dataset, visualized in a t-SNE embedding of the input data. B) The same trajectory visualized in an MDS embedding of GAP distances computed from the $RF_{uni}$ outlier detection model. C) Outlier scores of each data point along the trajectory from outlier to inliers. The corresponding image for each data point is shown, and its ground-truth anomaly label indicated by marker color. D) Visualization of feature importance of each pixel for the $RF_{uni}$ model, as a tally of how many partitions target each pixel. E) Counterfactual explanation of an outlier, showing the number of $RF_{uni}$ model partitions crossed (due to either increases or decreases in pixel values) to move directly from the first image to the last image in the trajectory. Blue indicates pixels where decreases in value explain reductions in outlier score, while red indicates where increases in the pixel value explain reductions in outlier score. F) Visualization of total $RF_{uni}$ partitions crossed, integrated over the counterfactual trajectory. Same color coding as in (E).}
	\label{Explainability_MNIST}
	\end{figure}

As we cannot meaningfully interpret explanations on the ADBench datasets (due to both preprocessing of the data, and required domain expertise), we turned to the MNIST dataset \citep{Deng2012}. For our experiment, we took a sample of the dataset comprising 90\% digit 9s (inliers) and 10\% digit 4s (anomalies). On this dataset, $RF_{uni}$ achieved an AUCROC of 0.76 (for reference, Isolation Forest achieved 0.75). Figure \ref{Explainability_MNIST}.a shows a trajectory computed for a digit 4, plotted in a t-SNE embedding of the original data \citep{vandermaaten08}. We also view this trajectory in an MDS embedding of GAP distances from the $RF_{uni}$ model (Fig. \ref{Explainability_MNIST}.b). The trajectory successfully identifies steps between similar observations in the MNIST dataset that reduce outlier score predictions (Fig. \ref{Explainability_MNIST}.c).

Figure \ref{Explainability_MNIST}.d visualizes the importance of each pixel to the $RF_{uni}$ model, as a function of how many Random Forest partitions target each pixel. Partitions crossed over the course of the counterfactual trajectory are then used to explain reductions in outlier score, either going directly from the first to the last observation (Fig. \ref{Explainability_MNIST}.e), or by integrating over the entire counterfactual trajectory (Fig. \ref{Explainability_MNIST}.f). Critically, these visualizations show not just differences in observations, but the salience of these differences to the $RF_{uni}$ anomaly detection model.

\section{Discussion}\label{sec:discussion}

The transformation that $RF_{uni}$ applies to data during similarity learning appears to be useful for anomaly detection. Compared to the $ExtraTrees$ model, which we show approximates a mapping of Euclidean distances, $RF_{uni}$ exhibits improved performance for the vast majority of benchmark datasets evaluated here. $RF_{uni}$ also compares very favorably to other unsupervised detectors such as KNN, Isolation Forest, local outlier factor, and PCA (Fig. \ref{ADbench_results}). The more popular approach of training an unsupervised Random Forest by generating synthetic data with the same marginal distribution, lacking its joint distribution (such as by shuffling features independently), performed significantly worse than $RF_{uni}$.

A notable property of supervised Random Forests is their insensitivity to coordinate transformations, monotone transformations of ordered features in the data (p.57 \citep{breiman1984classification}). This property is preserved for unsupervised learning in the Addcl1 case, where synthetic data are generated with the same marginal distribution as the real data, but lacking any joint distribution. This is because coordinate transformations applied to real data will be carried over and reflected in the synthetic data. Reducing feature values to effective rankings in this manner may be deleterious in many anomaly detection applications. The $RF_{uni}$ model uses the Addcl2 method for synthetic data generation, where the Random Forest is trained to discriminate between real data and synthetic data generated from a uniform distribution over the real data bounds. The resultant model is therefore sensitive to the relative values of features, not just their order, as the volume of synthetic data generated between real observations will be proportional to the difference in their relative values. While this is likely to be advantageous in many applications, some applications may benefit from pre-processing data with coordinate transformations, as shown to be the case for other unsupervised anomaly detection methods \citep{Kasieczka2023}.

Another attractive feature of $RF_{uni}$ is that, using a Random Forest, it can readily be applied to datasets of different modalities. This may include features of the measurement space, but also (and in tandem) engineered features, such as those from a convolutional neural network (CNN) for image datasets. For example, while $RF_{uni}$ achieved an AUCROC of 0.76 on the 4s and 9s subset of MNIST, using the bottleneck features from a CNN autoencoder improved performance to an AUCROC of 0.86\footnote{For this experiment we trained a two-layer Keras CNN autoencoder \citep{chollet2015keras}. The encoder consisted of a Conv2D layer with 32 3x3 pixel filters, a 2x2 pixel MaxPooling2D layer, a second Conv2D layer with 32 3x3 pixel filters, and a second 2x2 pixel MaxPooling2D layer. The decoder consisted of two Conv2DTranspose layers, each with 32 3x3 pixel filters and a stride of two, and a Conv2D layer with a single filter of 3x3 pixel convolution window.}.

In critical applications, model explainability is of particular value. Explainability improves transparency and trustworthiness, can safeguard against unfair and discriminatory processes, and provides actionable insights for users. $RF_{uni}$, by virtue of using an ensemble of decision trees, occupies a sweet spot between more expressive models such as neural networks, and more interpretable models such as multivariate regression. As we show here, feature importance can be read out both in terms of overall model functioning (Fig. \ref{Explainability_MNIST}.d) and to make sense of counterfactual explanations (Fig. \ref{Explainability_MNIST}.e,f).

A limitation of the $RF_{uni}$ model is that contamination of the training set with anomalies can significantly reduce performance, particularly if anomalies conform to a self-similar class-like distribution. However, this is a limitation for all unsupervised anomaly detection methods, and the reason we restrict experiments to datasets within ADBench where unsupervised detectors are appropriate. Although we do not explore it here, $RF_{uni}$ could be sensitive to outliers, as extreme values will determine the bounds over which the synthetic data is uniformly generated. In such cases, the majority of synthetic data may be situated far away from the real data, reducing the resolution with which $RF_{uni}$ can define the data manifold. While this can be mitigated by increasing the volume of synthetic data generated for training, a more robust approach would be to use not the bounds of the real data, but rather upper and lower percentiles.

\section{Conclusion}\label{sec:conclusion}

We describe an approach for anomaly detection using an unsupervised Random Forest for similarity learning. While similarity learning with unsupervised Random Forests is not new, here we focus on using a uniform distribution to generate the synthetic data that must be discriminated from the real data during training. Through analysis and visualizations we show that the resultant model anisometrically reshapes the data from measurement space, by expanding inter-point distances at the boundary of the data manifold. We show that this learned representation, and the exaggerated isolation of outliers in it, is particularly useful for anomaly detection, evaluated over a large collection of benchmark datasets. Finally, we demonstrate how predictions of this model can be rendered locally explainable, by interpreting counterfactuals through the lens of Random Forest feature importance.

\section{Appendix}

\subsection{ADBench datasets excluded from analysis}\label{sec:exclusions}

Datasets were excluded due to insufficient size (\emph{\# Samples}), over-represented anomalies (\emph{\% Anomalies}), or a dataset specific reason (\emph{Dataset}) \footnote{\textit{Optdigits}: Images of the digit zero vs nine other digits, which would not be naturally considered as `inliers'. \textit{ALOI}: Almost all unsupervised detectors performed below chance on this dataset, despite good performance of unsupervised detectors in prior literature \citep{Kriegel2011}.}.

\begin{table}[h]
\begin{center}
	\begin{tabular}{ l r r r|c c c c c c } 
	\hline
		Dataset & \# Samples & \# Features & \% Anomalies & COPOD & IForest & KNN & LOF & PCA & $RF_{uni}$ \\
		\hline
		\emph{ALOI} & 49534 & 27 & 3.04 &				0.50 & 0.49 & 0.50 & \textbf{0.58} & \textbf{0.51} & 0.51 \\ 
		Breastw & \emph{683} & 9 & \emph{34.99} &		0.99 & 0.99 & 0.98 & 0.38 & 0.96 & 0.99 \\ 
		Cardiotocography & 2114 & 21 & \emph{22.04} &	0.67 & 0.71 & 0.61 & 0.62 & 0.75 & 0.69 \\
		Fault & 1941 & 27 & \emph{34.67} & 				0.45 & 0.57 & \textbf{0.73} & 0.60 & 0.48 & 0.46 \\
		Glass & \emph{214} & 7 & 4.21 & 				0.76 & 0.79 & 0.86 & 0.81 & 0.69 & 0.80 \\
		Hepatitis & \emph{80} & 19 & 16.25 & 			\textbf{0.80} & 0.70 & 0.55 & 0.59 & 0.75 & \textbf{0.80} \\
		InternetAds & 1966 & 1555 & \emph{18.72} &		\textbf{0.67} & \textbf{0.68} & \textbf{0.69} & \textbf{0.65} & 0.62 & 0.56 \\
		Ionosphere & \emph{351} & 32 & \emph{35.90} &	0.79 & 0.85 & \textbf{0.93} & 0.86 & 0.78 & 0.84 \\
		Landsat & 6435 & 36 & \emph{20.71} &			0.43 & 0.49 & \textbf{0.62} & 0.54 & 0.38 & 0.56 \\
		Lymphography & \emph{148} & 18 & 4.05 &			1.00 & \textbf{1.00} & \textbf{1.00} & 0.98 & 1.00 & 0.98 \\
		Magic.gamma & 19020 & 10 & \emph{35.16} & 		0.67 & 0.72 & \textbf{0.77} & 0.74 & 0.66 & 0.69 \\
		\emph{Optdigits} & 5216 & 64 & 2.88 & 			0.70 & 0.65 & 0.50 & 0.39 & 0.52 & 0.45 \\
		Pima & \emph{768} & 8 & \emph{34.90} & 			0.65 & 0.67 & 0.62 & 0.54 & 0.63 & 0.66 \\
		Skin & 245057 & 3 & \emph{20.75} & 				0.50 & 0.67 & \textbf{0.72} & 0.39 & 0.44 & 0.35 \\
		SpamBase & 4207 & 57 & \emph{39.91} &  			0.70 & 0.63 & \textbf{0.73} & 0.57 & 0.55 & 0.49 \\
		Stamps & \emph{340} & 9 & 9.12 & 				\textbf{0.93} & 0.88 & 0.82 & 0.69 & 0.90 & 0.87 \\
		Vertebral & \emph{240} & 6 & 12.50 & 			0.33 & 0.36 & 0.33 & 0.49 & 0.38 & 0.29 \\
		WBC & \emph{223} & 9 & 4.48 & 					\textbf{0.99} & \textbf{1.00} & 0.99 & 0.83 & 0.99 & 0.99 \\
		WDBC & \emph{367} & 30 & 2.72 & 				0.99 & 0.99 & 1.00 & 1.00 & 0.99 & 0.98 \\
		Wine & \emph{129} & 13 & 7.75 &  				0.87 & 0.78 & 1.00 & \textbf{1.00} & 0.82 & 0.93 \\
		WPBC & \emph{198} & 33 & \emph{23.74} &  		0.52 & 0.49 & 0.52 & 0.52 & 0.48 & \textbf{0.57} \\
		Yeast & 1484 & 8 & \emph{34.16} & 				0.38 & 0.40 & 0.40 & \textbf{0.46} & 0.42 & 0.39 \\
	\hline
	\smallskip
	\end{tabular}
	\caption{\textbf{Anomaly detection performance for ADBench datasets excluded from analysis.} AUCROC for top performing unsupervised detectors. Bold values show dominance over other unsupervised detectors.}
	\label{Excluded_results}

	\end{center}
\end{table}


\subsection{Identifying Random Forest partition intersections for counterfactual trajectories}\label{sec:intersections}

To interpret counterfactual trajectory explanations in relation to $RF_{uni}$ feature importance, we identify which partitions of the Random Forest are intersected by the line segment connecting points in the trajectory. This is solved by an extension of the Liang–Barsky line clipping algorithm to higher dimensions \citep{Liang1984}.

For points $x_i$ and $x_{i+1}$ in the counterfactual trajectory, we first parameterize the line segment connecting them as $L$:

\[
	L = \{x(t) = x_i + t(x_{i+1} - x_i) \mid t \in [0,1]\} \subset \mathbb{R}^n
\]

Considering Random Forest partitions as a set of hyperrectangles, $R = \{R_1,R_2,...,R_m\}$, we want to find $R^*$, the subset of intersected partitions:

\[	
	R^* = \{R_j \in R \mid L \cap R_j \neq \emptyset \}
\]

Each partition hyperrectangle is defined by:

\[
	R_j = \{x \in X \mid \forall k \in \{1,\ldots,n\}: a_j^k \leq x^k \leq b_j^k\}
\]

Where $a_j^k$ and $b_j^k$ are the $k$-th coordinates of the minimum and maximum corners of $R_j$, respectively, for each dimension $k$ of $X \subseteq \mathbb{R}^n$. We can immediately filter for hyperrectangles whose partition feature value, $a^f=b^f$, is crossed by $L$:

\[
	R' = \{R_j \in R \mid a_j^f = b_j^f \implies \min(x_i^f, x_{i+1}^f) \leq a_j^f \leq \max(x_i^f, x_{i+1}^f)\}
\]

For remaining cases, we find the values of $t$ where the line segment would intersect the hyperrectangle bounds:

\[
	t_a^k = (a_j^k - x_i^k) / (x_{i+1}^k - x_i^k), \; t_b^k = (b_j^k - x_i^k) / (x_{i+1}^k - x_i^k)
\]

The intervals of $t$ where the segment lies within the bounds for each dimension $k$ are given by:

\[
	t_{min}^k = \min(t_a^k,t_b^k),\;t_{max}^k = \max(t_a^k,t_b^k)
\]

A partition intersection occurs if and only if the maximal value of $t_{min}$ is less than or equal to the minimal value of $t_{max}$ for all dimensions, and these values fall within $[0,1]$:

\[
	L \cap R' \neq \emptyset \iff \max_{k=1}^n(t_{min}^k) \leq \min_{k=1}^n(t_{max}^k) \; \land \; [\max_{k=1}^n(t_{min}^k), \min_{k=1}^n(t_{max}^k)] \cap [0,1] \neq \emptyset
\]	

The importance of each feature for a counterfactual is then given as $|R_k^*|$, the cardinality of the subset of hyperrectangles that take a single value for that dimension:

\[
	R_k^* = \{R_j \in R^* \mid a_j^k = b_j^k\}
\]

Feature importance can be further broken down with respect to the partition criteria. For numerical features, this can simply be a readout of whether partitions are intersected due to an increase or decrease in that feature, as shown in Figure \ref{Explainability_MNIST}.e-f.

\bibliographystyle{abbrvnat}
\bibliography{references}  

\begin{thebibliography}{46}
\providecommand{\natexlab}[1]{#1}
\providecommand{\url}[1]{\texttt{#1}}
\expandafter\ifx\csname urlstyle\endcsname\relax
  \providecommand{\doi}[1]{doi: #1}\else
  \providecommand{\doi}{doi: \begingroup \urlstyle{rm}\Url}\fi

\bibitem[Auret and Aldrich(2010)]{auret2010unsupervised}
L.~Auret and C.~Aldrich.
\newblock Unsupervised process fault detection with random forests.
\newblock \emph{Industrial \& Engineering Chemistry Research}, 49\penalty0 (19):\penalty0 9184--9194, 10 2010.
\newblock \doi{10.1021/ie901975c}.

\bibitem[Baron and Poznanski(2016)]{baron2016weirdest}
D.~Baron and D.~Poznanski.
\newblock The weirdest sdss galaxies: results from an outlier detection algorithm.
\newblock \emph{Monthly Notices of the Royal Astronomical Society}, 465\penalty0 (4):\penalty0 4530--4555, 11 2016.
\newblock ISSN 0035-8711.
\newblock \doi{10.1093/mnras/stw3021}.

\bibitem[Breiman and Cutler()]{breiman}
L.~Breiman and A.~Cutler.
\newblock "random forests".
\newblock \url{https://www.stat.berkeley.edu/~breiman/RandomForests/cc_home.htm#prox}.
\newblock Accessed: 2024-05-27.

\bibitem[Breiman and Cutler(2001)]{breiman-cutler-blog}
L.~Breiman and A.~Cutler.
\newblock Random forests.
\newblock \url{https://www.stat.berkeley.edu/~breiman/RandomForests/cc_home.htm}, 2001.

\bibitem[Breiman et~al.(1984)Breiman, Friedman, Stone, and Olshen]{breiman1984classification}
L.~Breiman, J.~Friedman, C.~Stone, and R.~Olshen.
\newblock \emph{Classification and Regression Trees}.
\newblock Taylor \& Francis, 1984.
\newblock ISBN 9780412048418.

\bibitem[Breunig et~al.(2000)Breunig, Kriegel, Ng, and Sander]{Breunig2000}
M.~M. Breunig, H.-P. Kriegel, R.~T. Ng, and J.~Sander.
\newblock Lof: identifying density-based local outliers.
\newblock \emph{SIGMOD Rec.}, 29\penalty0 (2):\penalty0 93–104, May 2000.
\newblock ISSN 0163-5808.
\newblock \doi{10.1145/335191.335388}.

\bibitem[Chollet et~al.(2015)]{chollet2015keras}
F.~Chollet et~al.
\newblock Keras.
\newblock \url{https://keras.io}, 2015.

\bibitem[Dem{\v{s}}ar(2006)]{demsar2006statistical}
J.~Dem{\v{s}}ar.
\newblock Statistical comparisons of classifiers over multiple data sets.
\newblock \emph{The Journal of Machine learning research}, 7\penalty0 (1):\penalty0 1--30, 2006.

\bibitem[Deng(2012)]{Deng2012}
L.~Deng.
\newblock The mnist database of handwritten digit images for machine learning research [best of the web].
\newblock \emph{IEEE Signal Processing Magazine}, 29\penalty0 (6):\penalty0 141--142, 2012.
\newblock \doi{10.1109/MSP.2012.2211477}.

\bibitem[Geurts et~al.(2006)Geurts, Ernst, and Wehenkel]{Geurts2006}
P.~Geurts, D.~Ernst, and L.~Wehenkel.
\newblock Extremely randomized trees.
\newblock \emph{Machine Learning}, 63\penalty0 (1):\penalty0 3--42, 2006.
\newblock \doi{10.1007/s10994-006-6226-1}.

\bibitem[Goldstein and Dengel(2012)]{Goldstein2012}
M.~Goldstein and A.~Dengel.
\newblock Histogram-based outlier score (hbos): A fast unsupervised anomaly detection algorithm.
\newblock 09 2012.

\bibitem[Gower(1966)]{gower1966}
J.~C. Gower.
\newblock Some distance properties of latent root and vector methods used in multivariate analysis.
\newblock \emph{Biometrika}, 53\penalty0 (3-4):\penalty0 325--338, 12 1966.
\newblock ISSN 0006-3444.
\newblock \doi{10.1093/biomet/53.3-4.325}.

\bibitem[Guidotti(2024)]{Guidotti2024}
R.~Guidotti.
\newblock Counterfactual explanations and how to find them: literature review and benchmarking.
\newblock \emph{Data Mining and Knowledge Discovery}, 38\penalty0 (5):\penalty0 2770--2824, 2024.
\newblock \doi{10.1007/s10618-022-00831-6}.

\bibitem[Han et~al.(2022)Han, Xiyang, Huang, Mingqi, and Zhao]{adbench}
S.~Han, H.~Xiyang, H.~Huang, J.~Mingqi, and Y.~Zhao.
\newblock Adbench: Anomaly detection benchmark.
\newblock \emph{Advances in Neural Information Processing Systems (NeurIPS 2022)}, 35:\penalty0 32142--32159, 2022.

\bibitem[He et~al.(2003)He, Xu, and Deng]{He2003}
Z.~He, X.~Xu, and S.~Deng.
\newblock Discovering cluster-based local outliers.
\newblock \emph{Pattern Recognition Letters}, 24\penalty0 (9):\penalty0 1641--1650, 2003.
\newblock ISSN 0167-8655.
\newblock \doi{https://doi.org/10.1016/S0167-8655(03)00003-5}.

\bibitem[Kasieczka et~al.(2023)Kasieczka, Mastandrea, Mikuni, Nachman, Pettee, and Shih]{Kasieczka2023}
G.~Kasieczka, R.~Mastandrea, V.~Mikuni, B.~Nachman, M.~Pettee, and D.~Shih.
\newblock Anomaly detection under coordinate transformations.
\newblock \emph{Phys. Rev. D}, 107:\penalty0 015009, Jan 2023.
\newblock \doi{10.1103/PhysRevD.107.015009}.

\bibitem[Kriegel et~al.()Kriegel, Kroger, Schubert, and Zimek]{Kriegel2011}
H.-P. Kriegel, P.~Kroger, E.~Schubert, and A.~Zimek.
\newblock \emph{Interpreting and Unifying Outlier Scores}, pages 13--24.
\newblock \doi{10.1137/1.9781611972818.2}.

\bibitem[Kriegel et~al.(2009)Kriegel, Kr{\"o}ger, Schubert, and Zimek]{Kriegel2009}
H.-P. Kriegel, P.~Kr{\"o}ger, E.~Schubert, and A.~Zimek.
\newblock Outlier detection in axis-parallel subspaces of high dimensional data.
\newblock In T.~Theeramunkong, B.~Kijsirikul, N.~Cercone, and T.-B. Ho, editors, \emph{Advances in Knowledge Discovery and Data Mining}, pages 831--838, Berlin, Heidelberg, 2009. Springer Berlin Heidelberg.
\newblock ISBN 978-3-642-01307-2.

\bibitem[Latecki et~al.(2007)Latecki, Lazarevic, and Pokrajac]{Latecki2007}
L.~J. Latecki, A.~Lazarevic, and D.~Pokrajac.
\newblock Outlier detection with kernel density functions.
\newblock In P.~Perner, editor, \emph{Machine Learning and Data Mining in Pattern Recognition}, pages 61--75, Berlin, Heidelberg, 2007. Springer Berlin Heidelberg.
\newblock ISBN 978-3-540-73499-4.

\bibitem[Li et~al.(2020)Li, Zhao, Botta, Ionescu, and Hu]{Li2020}
Z.~Li, Y.~Zhao, N.~Botta, C.~Ionescu, and X.~Hu.
\newblock { COPOD: Copula-Based Outlier Detection }.
\newblock In \emph{2020 IEEE International Conference on Data Mining (ICDM)}, pages 1118--1123, Los Alamitos, CA, USA, Nov. 2020. IEEE Computer Society.
\newblock \doi{10.1109/ICDM50108.2020.00135}.

\bibitem[Li et~al.(2023)Li, Zhao, Hu, Botta, Ionescu, and Chen]{Li2023}
Z.~Li, Y.~Zhao, X.~Hu, N.~Botta, C.~Ionescu, and G.~H. Chen.
\newblock Ecod: Unsupervised outlier detection using empirical cumulative distribution functions.
\newblock \emph{IEEE Transactions on Knowledge and Data Engineering}, 35\penalty0 (12):\penalty0 12181--12193, 2023.
\newblock \doi{10.1109/TKDE.2022.3159580}.

\bibitem[Liang and Barsky(1984)]{Liang1984}
Y.-D. Liang and B.~A. Barsky.
\newblock A new concept and method for line clipping.
\newblock \emph{ACM Trans. Graph.}, 3\penalty0 (1):\penalty0 1–22, Jan. 1984.
\newblock ISSN 0730-0301.
\newblock \doi{10.1145/357332.357333}.

\bibitem[Liu et~al.(2008)Liu, Ting, and Zhou]{Liu2008}
F.~T. Liu, K.~M. Ting, and Z.-H. Zhou.
\newblock Isolation forest.
\newblock In \emph{2008 Eighth IEEE International Conference on Data Mining}, pages 413--422, 2008.
\newblock \doi{10.1109/ICDM.2008.17}.

\bibitem[Lundberg and Lee(2017)]{Lundberg2017}
S.~M. Lundberg and S.-I. Lee.
\newblock A unified approach to interpreting model predictions.
\newblock In I.~Guyon, U.~V. Luxburg, S.~Bengio, H.~Wallach, R.~Fergus, S.~Vishwanathan, and R.~Garnett, editors, \emph{Advances in Neural Information Processing Systems}, volume~30. Curran Associates, Inc., 2017.

\bibitem[Madhyastha et~al.(2019)Madhyastha, Li, Browne, Strnadova-Neeley, Priebe, Burns, and Vogelstein]{madhyastha2019geodesic}
M.~Madhyastha, P.~Li, J.~Browne, V.~Strnadova-Neeley, C.~E. Priebe, R.~Burns, and J.~T. Vogelstein.
\newblock Geodesic learning via unsupervised decision forests, 2019.

\bibitem[Mensi et~al.(2022)Mensi, Cicalese, and Bicego]{mensi2022rf}
A.~Mensi, F.~Cicalese, and M.~Bicego.
\newblock \emph{Using Random Forest Distances for Outlier Detection}, pages 75--86.
\newblock 05 2022.
\newblock ISBN 978-3-031-06432-6.
\newblock \doi{10.1007/978-3-031-06433-3_7}.

\bibitem[Munir et~al.(2019{\natexlab{a}})Munir, Siddiqui, Chattha, Dengel, and Ahmed]{munirFuseAD}
M.~Munir, S.~A. Siddiqui, M.~A. Chattha, A.~Dengel, and S.~Ahmed.
\newblock Fusead: Unsupervised anomaly detection in streaming sensors data by fusing statistical and deep learning models.
\newblock \emph{Sensors}, \penalty0 (11):\penalty0 1991--2005, 2019{\natexlab{a}}.

\bibitem[Munir et~al.(2019{\natexlab{b}})Munir, Siddiqui, Dengel, and Ahmed]{munirDeepAnT}
M.~Munir, S.~A. Siddiqui, A.~Dengel, and S.~Ahmed.
\newblock Deepant: A deep learning approach for unsupervised anomaly detection in time series.
\newblock \emph{IEEE Access}, 7:\penalty0 1991--2005, 2019{\natexlab{b}}.
\newblock \doi{10.1109/ACCESS.2018.2886457}.

\bibitem[Novello et~al.(2024)Novello, Dalmau, and Andeol]{novello2024outofdistributiondetectionuseconformal}
P.~Novello, J.~Dalmau, and L.~Andeol.
\newblock Out-of-distribution detection should use conformal prediction (and vice-versa?), 2024.

\bibitem[Olteanu et~al.(2023)Olteanu, Rossi, and Yger]{olteanu}
M.~Olteanu, F.~Rossi, and F.~Yger.
\newblock Meta-survey on outlier and anomaly detection.
\newblock \emph{Neurocomputing}, 555, 2023.

\bibitem[Pevn{\'y}(2016)]{Pevny2016}
T.~Pevn{\'y}.
\newblock Loda: Lightweight on-line detector of anomalies.
\newblock \emph{Machine Learning}, 102\penalty0 (2):\penalty0 275--304, 2016.
\newblock \doi{10.1007/s10994-015-5521-0}.

\bibitem[Puggini et~al.(2015)Puggini, Doyle, and McLoone]{PUGGINI2015583}
L.~Puggini, J.~Doyle, and S.~McLoone.
\newblock Fault detection using random forest similarity distance.
\newblock \emph{IFAC-PapersOnLine}, 48\penalty0 (21):\penalty0 583--588, 2015.
\newblock ISSN 2405-8963.
\newblock \doi{https://doi.org/10.1016/j.ifacol.2015.09.589}.
\newblock 9th IFAC Symposium on Fault Detection, Supervision and Safety for Technical Processes SAFEPROCESS 2015.

\bibitem[Ramaswamy et~al.(2000)Ramaswamy, Rastogi, and Shim]{Ramaswamy2000}
S.~Ramaswamy, R.~Rastogi, and K.~Shim.
\newblock Efficient algorithms for mining outliers from large data sets.
\newblock \emph{SIGMOD Rec.}, 29\penalty0 (2):\penalty0 427–438, May 2000.
\newblock ISSN 0163-5808.
\newblock \doi{10.1145/335191.335437}.

\bibitem[Rhodes et~al.(2023{\natexlab{a}})Rhodes, Cutler, and Moon]{rhodes}
J.~S. Rhodes, A.~Cutler, and K.~R. Moon.
\newblock Geometry- and accuracy-preserving random forest proximities.
\newblock \emph{IEEE Transactions on Pattern Analysis and Machine Intelligence}, 45\penalty0 (9):\penalty0 10947--10959, 2023{\natexlab{a}}.

\bibitem[Rhodes et~al.(2023{\natexlab{b}})Rhodes, Cutler, and Moon]{rhodes2023geometry}
J.~S. Rhodes, A.~Cutler, and K.~R. Moon.
\newblock Geometry- and accuracy-preserving random forest proximities, 2023{\natexlab{b}}.

\bibitem[Ribeiro et~al.(2016)Ribeiro, Singh, and Guestrin]{Ribeiro2016}
M.~T. Ribeiro, S.~Singh, and C.~Guestrin.
\newblock "why should i trust you?": Explaining the predictions of any classifier.
\newblock In \emph{Proceedings of the 22nd ACM SIGKDD International Conference on Knowledge Discovery and Data Mining}, KDD '16, page 1135–1144, New York, NY, USA, 2016. Association for Computing Machinery.
\newblock ISBN 9781450342322.
\newblock \doi{10.1145/2939672.2939778}.

\bibitem[Santini and Jain(1999)]{santini}
S.~Santini and R.~Jain.
\newblock Similarity measures.
\newblock \emph{IEEE Transactions on Pattern Analysis and Machine Intelligence}, 21\penalty0 (9):\penalty0 871--883, 1999.

\bibitem[Sch\"{o}lkopf et~al.(1999)Sch\"{o}lkopf, Williamson, Smola, Shawe-Taylor, and Platt]{Scholkopf1999}
B.~Sch\"{o}lkopf, R.~C. Williamson, A.~Smola, J.~Shawe-Taylor, and J.~Platt.
\newblock Support vector method for novelty detection.
\newblock In S.~Solla, T.~Leen, and K.~M\"{u}ller, editors, \emph{Advances in Neural Information Processing Systems}, volume~12. MIT Press, 1999.

\bibitem[Seligson et~al.(2005)Seligson, Horvath, Shi, Yu, Tze, Grunstein, and Kurdistani]{seligson2005global}
D.~B. Seligson, S.~Horvath, T.~Shi, H.~Yu, S.~Tze, M.~Grunstein, and S.~K. Kurdistani.
\newblock Global histone modification patterns predict risk of prostate cancer recurrence.
\newblock \emph{Nature}, 435\penalty0 (7046):\penalty0 1262--1266, 2005.
\newblock \doi{10.1038/nature03672}.

\bibitem[Shi and Horvath(2006)]{shi}
T.~Shi and S.~Horvath.
\newblock Unsupervised learning with random forest predictors.
\newblock \emph{Journal of Computational and Graphical Statistics}, 15\penalty0 (1):\penalty0 118--138, 2006.

\bibitem[Shyu et~al.(2003)Shyu, Chen, Sarinnapakorn, and Chang]{Shyu2003}
M.-L. Shyu, S.-C. Chen, K.~Sarinnapakorn, and L.~Chang.
\newblock A novel anomaly detection scheme based on principal component classifier.
\newblock 01 2003.

\bibitem[Simonyan et~al.(2013)Simonyan, Vedaldi, and Zisserman]{simonyan2013deep}
K.~Simonyan, A.~Vedaldi, and A.~Zisserman.
\newblock Deep inside convolutional networks: Visualising image classification models and saliency maps.
\newblock \emph{arXiv preprint arXiv:1312.6034}, 2013.

\bibitem[Tang et~al.(2002)Tang, Chen, Fu, and Cheung]{Tang2002}
J.~Tang, Z.~Chen, A.~W.-c. Fu, and D.~W. Cheung.
\newblock Enhancing effectiveness of outlier detections for low density patterns.
\newblock In M.-S. Chen, P.~S. Yu, and B.~Liu, editors, \emph{Advances in Knowledge Discovery and Data Mining}, pages 535--548, Berlin, Heidelberg, 2002. Springer Berlin Heidelberg.
\newblock ISBN 978-3-540-47887-4.

\bibitem[Tversky(1977)]{tversky}
A.~Tversky.
\newblock Features of similarity.
\newblock \emph{Psychological Review}, 84\penalty0 (4):\penalty0 327--352, 1977.

\bibitem[van~der Maaten and Hinton(2008)]{vandermaaten08}
L.~van~der Maaten and G.~Hinton.
\newblock Visualizing data using t-sne.
\newblock \emph{Journal of Machine Learning Research}, 9\penalty0 (86):\penalty0 2579--2605, 2008.

\bibitem[Zhao et~al.(2019)Zhao, Zain, and Li]{pyod}
Y.~Zhao, N.~Zain, and Z.~Li.
\newblock Pyod: A python toolbox for scalable outlier detection.
\newblock \emph{Journal of Machine Learning Research}, pages 1--7, 2019.

\end{thebibliography}






\end{document}